\documentclass[english]{article}
\usepackage[latin9]{inputenc}
\usepackage{geometry}
\usepackage{amsmath}
\usepackage{amssymb}
\usepackage{graphicx}

\makeatletter

\providecommand{\tabularnewline}{\\}

\@ifundefined{date}{}{\date{}}

\usepackage{array}\usepackage{float}\usepackage{epstopdf}

\usepackage{units}

\providecommand{\tabularnewline}{\\}
\usepackage{babel}

\renewcommand{\v}{\mathbf{v}}
\newcommand{\x}{\mathbf{x}}
\renewcommand{\o}{\mathbf{o}}
\newcommand{\e}{\mathbf{e}}
\newcommand{\h}{\mathbf{h}}

\makeatother

\usepackage{babel}
\begin{document}

\title{MCMC for Hierarchical Semi-Markov Conditional Random Fields}

\author{Tran The Truyen, Dinh Q. Phung, Svetha Venkatesh\\
 Department of Computing, Curtin University of Technology, Australia\\
 \texttt{\{t.tran2,d.phung,s.venkatesh\}@curtin.edu.au}\\
 \\
 Hung H. Bui\\
 Artificial Intelligence Center, SRI International, USA\\
 \texttt{bui@ai.sri.com}}
\maketitle
\begin{abstract}
Deep architecture such as hierarchical semi-Markov models is an important
class of models for nested sequential data. Current exact inference
schemes either cost cubic time in sequence length, or exponential
time in model depth. These costs are prohibitive for large-scale problems
with arbitrary length and depth. In this contribution, we propose
a new approximation technique that may have the potential to achieve
sub-cubic time complexity in length and linear time depth, at the
cost of some loss of quality. The idea is based on two well-known
methods: Gibbs sampling and Rao-Blackwellisation. We provide some
simulation-based evaluation of the quality of the RGBS with respect
to run time and sequence length. 
\end{abstract}

\section{Introduction}

Hierarchical Semi-Markov models such as HHMM \cite{fine98hierarchical}
and HSCRF \cite{Truyen:2008a} are deep generalisations of the HMM
\cite{RabinerIEEE-89} and the linear-chain CRF \cite{lafferty01conditional},
respectively. These models are suitable for data that follows nested
Markovian processes, in that a state in a sub-Markov chain is also
a Markov chain at the child level. Thus, in theory, we can model arbitrary
depth of semantics for sequential data. The models are essentially
members of the Probabilistic Context-Free Grammar family with bounded
depth.

However, the main drawback of these formulations is that the inference
complexity, as inherited from the Inside-Outside algorithm of the
Context-Free Grammars, is cubic in sequence length. As a result, this
technique is only appropriate for short data sequences, e.g. in NLP
we often need to limit the sentence length to, says, $30$. There
exists a linearisation technique proposed in \cite{Murphy-Paskin01},
in that the HHMM is represented as a Dynamic Bayesian Network (DBN).
By collapsing all states within each time slice of the DBN, we are
able achieve linear complexity in sequence length, but exponential
complexity in depth. Thus, this technique cannot handle deep architectures.

In this contribution, we introduce an approximation technique using
Gibbs samplers that have a potential of achieving sub-cubic time complexity
in sequence length and linear time in model depth. The idea is that,
although the models are complex, the nested property allows only one
state transition at a time across all levels. Secondly, if all the
state transitions are known, then the model can be collapsed into
a Markov tree, which is efficient to evaluate. Thus the trick is to
sample only the Markov transition at each time step and integrating
over the state variables. This trick is known as Rao-Blackwellisation,
which has previously been applied for DBNs \cite{doucet2000rao}.
Thus, we call this method Rao-Blackwellisation Gibbs Sampling (RBGS).
Of course, as a MCMC method, the price we have to pay is some degradation
in inference quality.

\section{Background}

\subsection{Hierarchical Semi-Markov Conditional Random Fields \label{sub:Hieararchical-Semi-Markov-Conditional}}

Recall that in the linear-chain CRFs \cite{lafferty01conditional},
we are given a sequence of observations $\o=(o_{1},o_{2},...,o_{T})$
and a corresponding sequence of state variables $\x=(x_{1},x_{2},...,x_{T})$.
The model distribution is then defined as 
\[
P(\x|\o)=\frac{1}{Z(\o)}\prod_{t=1}^{T-1}\phi_{t}(x_{t},x_{t+1},\o)
\]
where $\phi_{t}(x_{t},x_{t+1},\o)>0$ are potential functions that
capture the association between $\o$ and $\{x_{t},x_{t+1}\}$ as
well as the \emph{transition} between state $x_{t}$ to state $x_{t+1}$,
and $Z(\o)$ is the normalisation constant.

Thus, given the observation $\o$, the model admits the Markovian
property in that $P(x_{t}|\x_{\neg t})=P(x_{t}|x_{t-1},x_{t+1})$,
where $\x_{\neg t}$ is a shorthand for $\x\backslash x_{t}$. This
is clearly a simplified assumption but it allows fast inference in
$\mathcal{O}(T)$ time, and more importantly, it has been widely proved
useful in practice. On the other hand, in some applications where
the state transitions are not strictly Markovian, i.e. the states
tend to be persistent for an extended time. A better way is to assume
only the transition between \emph{parent-states} $s=(x_{t},x_{t+1},...,x_{t+l})$,
whose elements are not necessarily Markovian. This is the idea behind
the semi-Markov model, which has been introduced in \cite{sarawagi04}
in the context of CRFs. The inference complexity of the semi-Markov
models is generally $\mathcal{O}(T^{2})$ since we have to account
for all possible segment lengths.

The Hierarchical Semi-Markov Conditional Random Field (HSCRF) is the
generalisation of the semi-Markov model in the way that the parent-state
is also an element of the grandparent-state at the higher level. In
effect, we have a \emph{fractal} sequential architecture, in that
there are multiple levels of detail, and if we examine one level,
it looks exactly like a Markov chain, but each state in the chain
is a sub-Markov chain at the lower level. This may capture some real-world
phenomena, for example, in NLP we have multiple levels such as character,
unigram, word, phrase, clause, sentence, paragraph, section, chapter
and book. The price we pay for these expressiveness is the increase
in inference complexity to $\mathcal{O}(T^{3})$.

One of the most important properties that we will exploit in this
contribution is the \emph{nestedness}, in that a parent can only transits
to a new parent if its child chain has terminated. Conversely, when
a child chain is still active, the parent state must stay the same.
For example, in text when a noun-phrase is said to transit to a verb-phrase,
the subsequence of words within the noun-phase must terminate, and
at the same time, the noun-phrase and the verb-phrase must belong
to the same clause.

\begin{figure}[htb]
\begin{centering}
\begin{tabular}{c}
\includegraphics[width=0.3\linewidth]{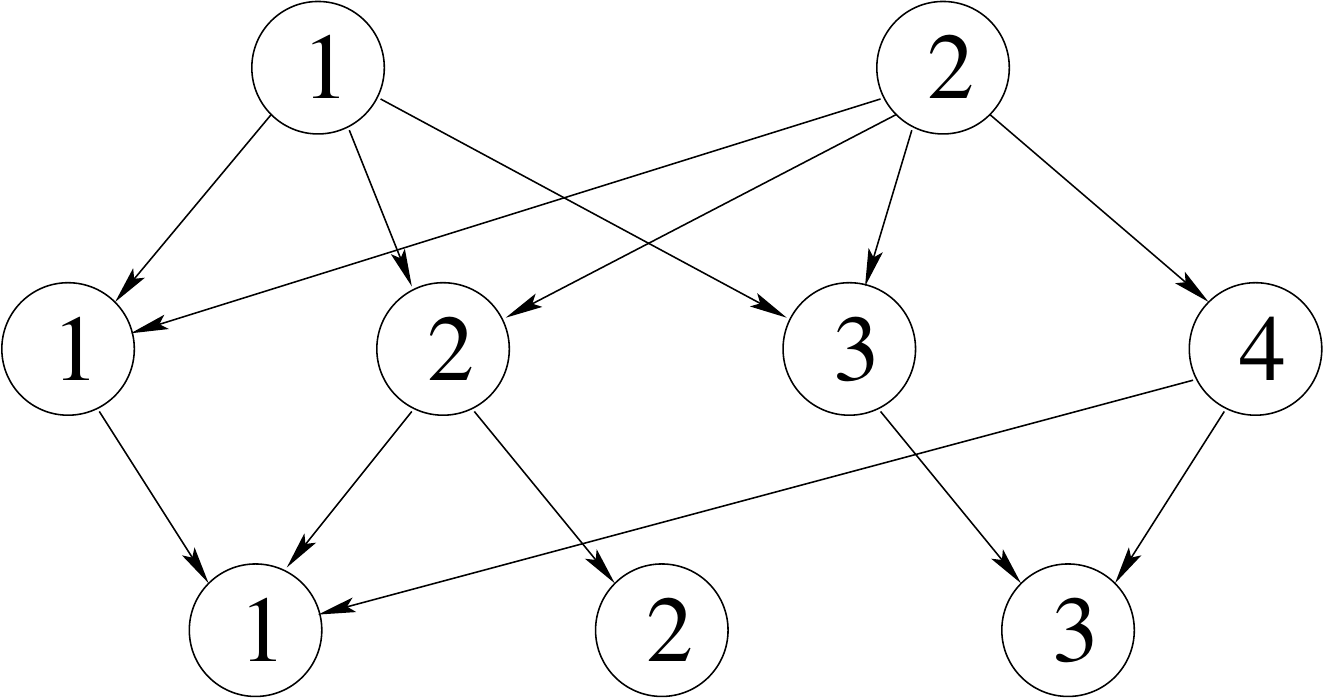} \tabularnewline
\tabularnewline
\end{tabular}
\par\end{centering}

\caption{State topological structure in HSCRFs.}

\label{fig:topo-model} 
\end{figure}

The parent-child relations in the HSCRF can be described using a state
hierarchical topology. Figure~\ref{fig:topo-model} depicts a three-level
topology, where the top, middle and bottom levels have two, four,
and three states respectively. Each child has multiple parents and
each parent may share the same subset of children. Note that, this
is already a generalisation over the topology proposed in the original
HHMM \cite{fine98hierarchical}, where each child has exactly one
parent.

\subsection{MCMC Methods}

In this subsection, we briefly review the two ideas that would lead
to our proposed approximation method: the Gibbs sampling (e.g. see~\cite{Geman-Geman84})
and Rao-Blackwellisation (e.g. see~\cite{casella1996rbs}). The main
idea behind Gibbs sampling is that, given a set of variables $\x=(x_{1},x_{2},...,x_{T})$
we can cyclically sample one variable at a time given the rest of
variables, i.e. 
\begin{eqnarray*}
\tilde{x}{}_{t} & \sim & P(x_{t}|\x_{\neg t})
\end{eqnarray*}
and this will eventually converge to the true distribution $P(\x)$.
This method is effective if the conditional distribution $P(x_{t}|\x_{\neg t})$
is easy to compute and sample from. The main drawback is that it can
take a long time to converge.

Rao-Blackwellisation is a technique that can improve the quality of
sampling methods by only sampling some variables and integrating over
the rest. Clearly, Rao-Blackwellisation is only possibly if the integration
is easy to perform analytically. Specifically, supposed that we have
the decomposition $\x=(\v,\h)$ and the marginalisation $P(\v)=\sum_{\h}P(\v,\h)$
can be evaluated efficiently for each $\v$, then according to the
Rao-Blackwell theorem, sampling $\v$ from $P(\v)$ would yield smaller
variance than sampling both $(\v,\h)$ from $P(\v,\h)$.

\section{Rao-Blackwellised Gibbs Sampling}

In the HSCRFs, we only specify the general topological structure of
the states, as well as the model depth $D$ and length $T$. The dynamics
of the models are then automatically realised by multiple \emph{events}:
(i) the the parent starts its life by \emph{emitting} a child, (ii)
the child, after finishing its role, \emph{transits} to a new child,
and (iii) when the child Markov chain \emph{terminates}, it returns
control to the parent. At the beginning of the sequence, the emission
is activated from the top down to the bottom. At the end of the sequence,
the termination first occurs at the bottom level and then continues
to the top.

The main complication is that since the dynamics is not known beforehand,
we have to account for every possible event when doing inference.
Fortunately, the nested property discussed in Section~\ref{sub:Hieararchical-Semi-Markov-Conditional}
has two important implications that would help to avoid explicitly
enumerating these exponentially many events. Firstly, for any model
of depth $D$, there is \emph{exactly one transition} occurring at
a specific time $t$. This is because, suppose the transition occurs
at level $d$ then all the states at $d'$ above it (i.e. $d'<d$)
must stay the same, and all the states at level $d''$ below it (i.e.
$d''>d$) must have already ended. Secondly, suppose that all the
transitions at any time $1\le t\le T$ are known then the whole model
can be collapsed into a Markov tree, which is efficient to evaluate.
These implications have more or less been exploited in a family of
algorithms known as \emph{Inside-Outside}, which has the complexity
of $\mathcal{O}(T^{3})$.

In this section, we consider the case that the sequence length $T$
is sufficiently large, e.g. $T\ge50$, then the cubic complexity is
too expensive for practical problems. Let us denote by $\e=(e_{1},e_{2},..,e_{T})$
the set of transitions, i.e. $e_{t}=d$ for some $1\le d\le D$. These
transition variables, together with the state variables $\x$ and
the observational $\o$ completely specify the variable configuration
of the model, which has the probability of

\[
P(\x,\e|\o)=\frac{1}{Z(\o)}\Phi(\x,\e,\o)
\]
where $\Phi(\x,\e,\o)$ is the joint potential function.

The idea of Rao-Blackwellised Gibbs Sampling is that, if we can efficiently
estimate the transition levels $\{e_{t}\}$ at any time $t$, then
the model evaluation using $P(\x|\e,\o)$ is feasible since the model
is now essentially a Markov tree. Thus, what remains is the estimation
of $\{e_{t}$\} from $P(\e|\o)=\sum_{\x}P(\x,\e|\o)$. It turns out
that although we cannot enumerate $P(\e|\o)$ for all $\e\in\{1,2,...,D\}^{T}$
directly, the following quantities can be inexpensive to compute 
\[
P(e_{t}|\e_{\neg t},\o)\propto\sum_{\x}P(e_{t},\x|\e_{\neg t},\o)
\]
This suggests that we can use the Gibbs sampler to obtain samples
of $\e$. This is Rao-Blackwellisation because of the integration
over $\x$, that is, we sample $\e$ without the need for sampling
$\x$.

It should be stressed that the straightforward implementation may
be expensive if we sum over $\x$ for every $e_{t}$ for $1\le t\le T$.
Fortunately, there is a better way, in that we proceed from left-to-right
using an idea known as \emph{walking-chain} \cite{Bui-et-al02}, which
is essentially a generalisation of the forward-backward procedure.
For space limitations, we omit the derivation and interested readers
may consult \cite[Ch. 9]{truyen-thesis08} for more detail. The overall
result is that we can obtain a full sample of $\{e_{t}\}$ for all
$t$ in $DT$ time.

We note in passing that the walking-chain was originally introduced
in the context of the Abstract Hidden Markov Model (AHMM), which does
not model the duration of the state persistence while the HSCRF does.
Secondly, as the HSCRF is undirected, its factorisation of potentials
does not have any probabilistic interpretation as in the AHMM.

\section{Evaluation}

\begin{figure}[htb]
\begin{centering}
\begin{tabular}{cc}
\includegraphics[width=0.4\linewidth]{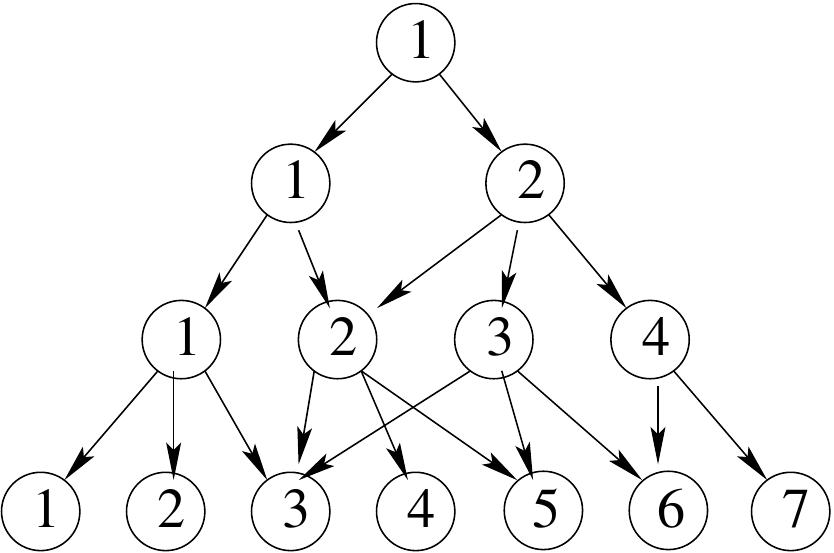}  & \tabularnewline
\end{tabular}
\par\end{centering}

\caption{Topology used in experiments}

\label{fig:topo-exp} 
\end{figure}

Using the HHMM simulator previously developed in \cite{Bui-et-al04}
we generate data according to some fixed parameters and topology (Figure~\ref{fig:topo-exp}).
Specifically, the model has the depth of $D=4$ and sequence length
of $T=30$, state size of $\{1,2,4,7\}$ for the four semantic levels,
respectively. At the bottom level, $3$ observation symbols that are
outputted in a generative manner. We generate $50$ training sequences
for learning and $50$ for testing.

First, we learn the HSCRF parameters from the training data. Note
that the HSCRF is discriminative, the learned model may not correspond
to the original HHMM that generates the data. Given the learned HSCRF,
we perform various inference tasks with the testing data. We compare
the results of Rao-Blackwellised Gibbs sampling introduced in this
paper against the exact inference using the Inside-Outside algorithm
described in \cite{Truyen:2008a}.

For sampling, first we run the sampler for a `burn-in' period of time,
and then discard those samples. The purpose of this practice is to
let the sampler `forget' about the starting point to eliminate some
possible bias. The burn-in time is set to about $10\%$ of the total
iterations.

We want to examine the accuracy of the proposed approximation against
the exact methods, along the three dimensions: (i) the state marginals
$P(x_{t}^{d}|\o)$ at time $t$ and depth level $d$, (ii) the probability
$P(e_{t}|\o)$ that a transition occurs at time $t$, and (iii) the
decoding accuracy using $\hat{x}{}_{t}^{d}=\arg\max_{x_{t}^{d}}P(x_{t}^{d}|\o)$.
For the first two estimations, two accuracy measures are used: a)
the average Kullback-Leibler divergence between state marginals and
b) the average absolute difference in state marginals. For decoding
evaluation, we use the percentage of matching between the maximal
states.

\begin{figure}[htb]
\begin{centering}
\begin{tabular}{cc}
\includegraphics[width=0.5\linewidth]{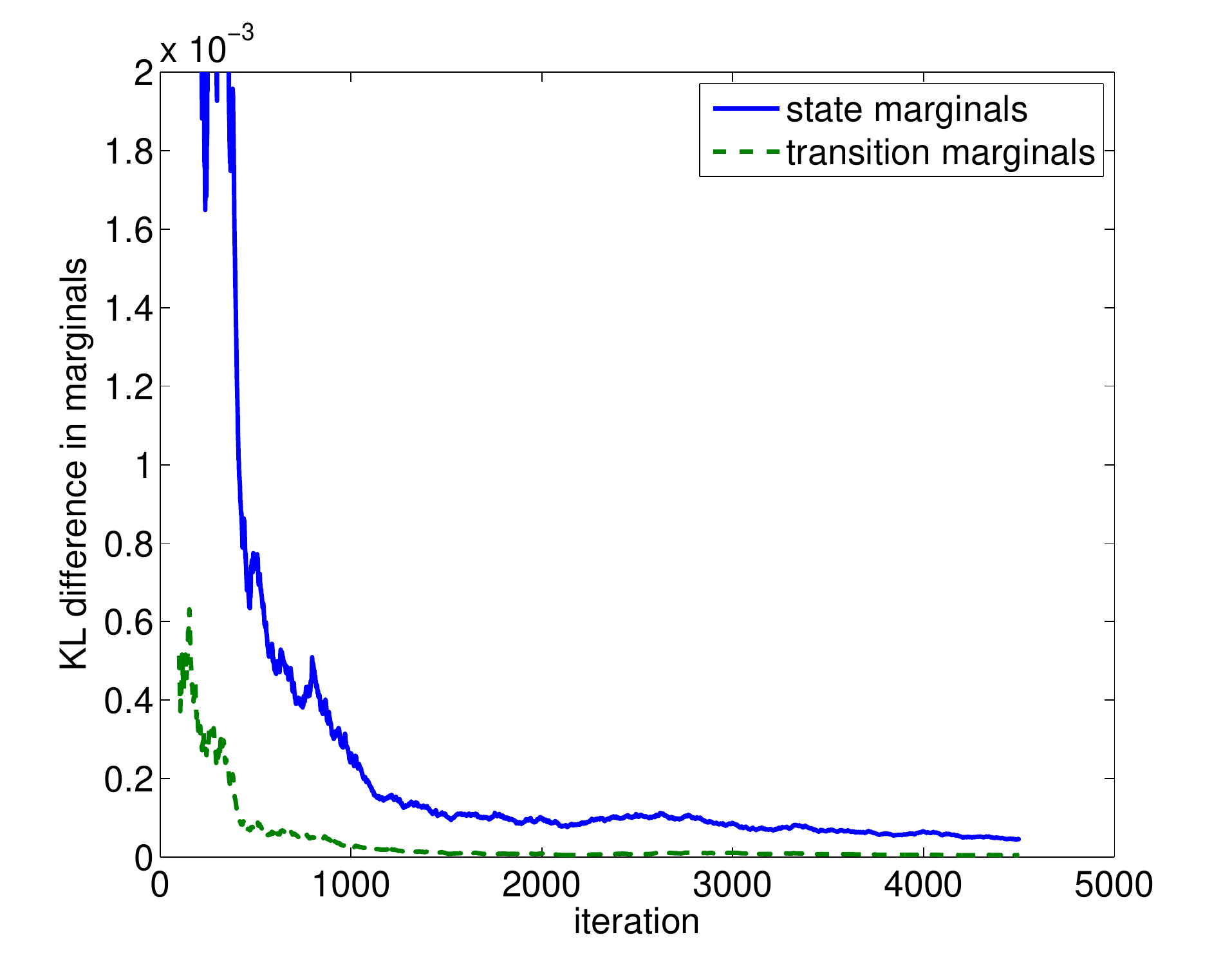}  & \includegraphics[width=0.5\linewidth]{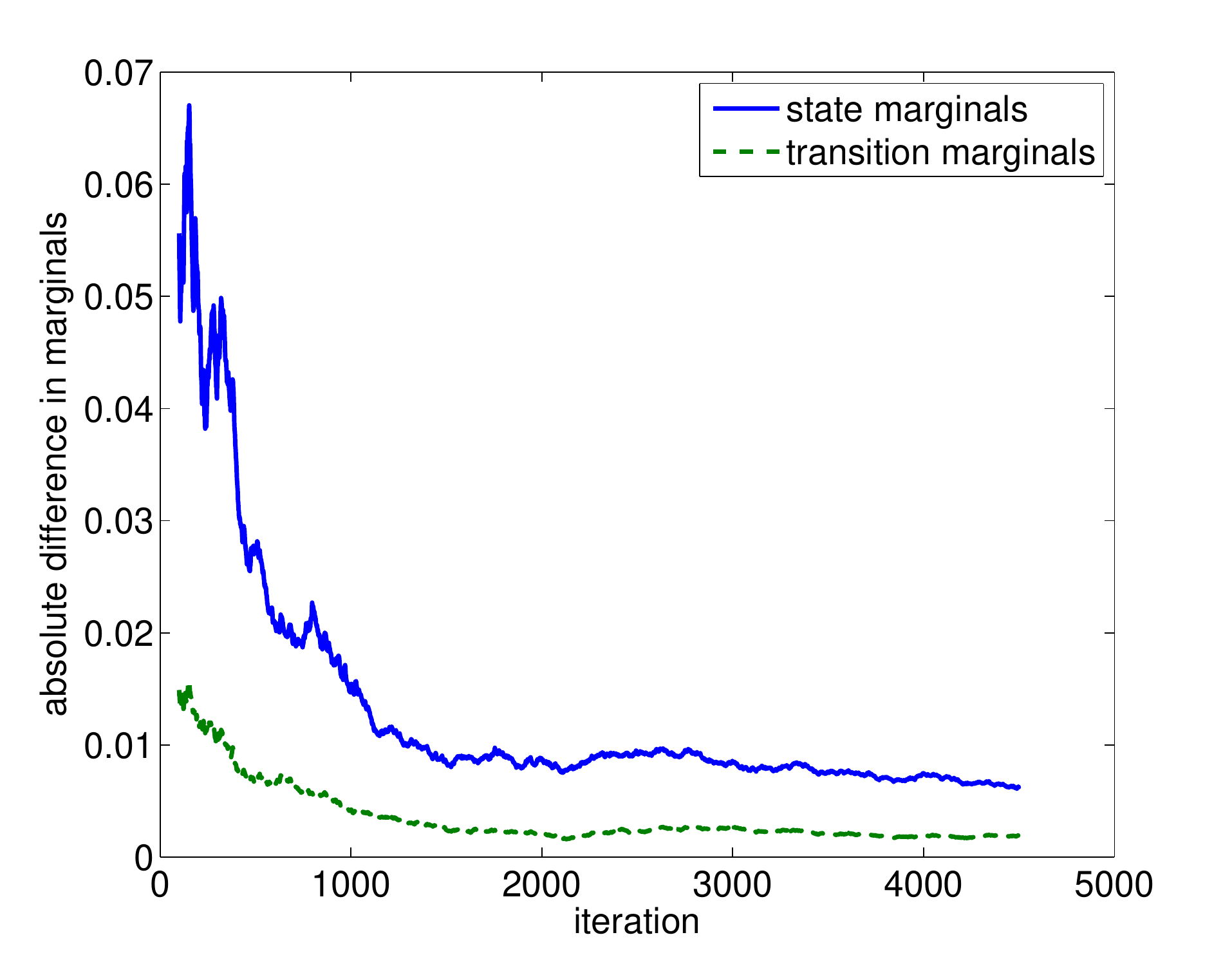}\tabularnewline
(a)  & (b)\tabularnewline
\end{tabular}
\par\end{centering}

\caption{Divergence between the Gibbs estimation and the true marginals: (a)
KL-divergence, and (b) $l_{1}$-norm difference.}

\label{fig:divergences} 
\end{figure}

\begin{figure}[htb]
\begin{centering}
\begin{tabular}{c}
\includegraphics[width=0.5\linewidth]{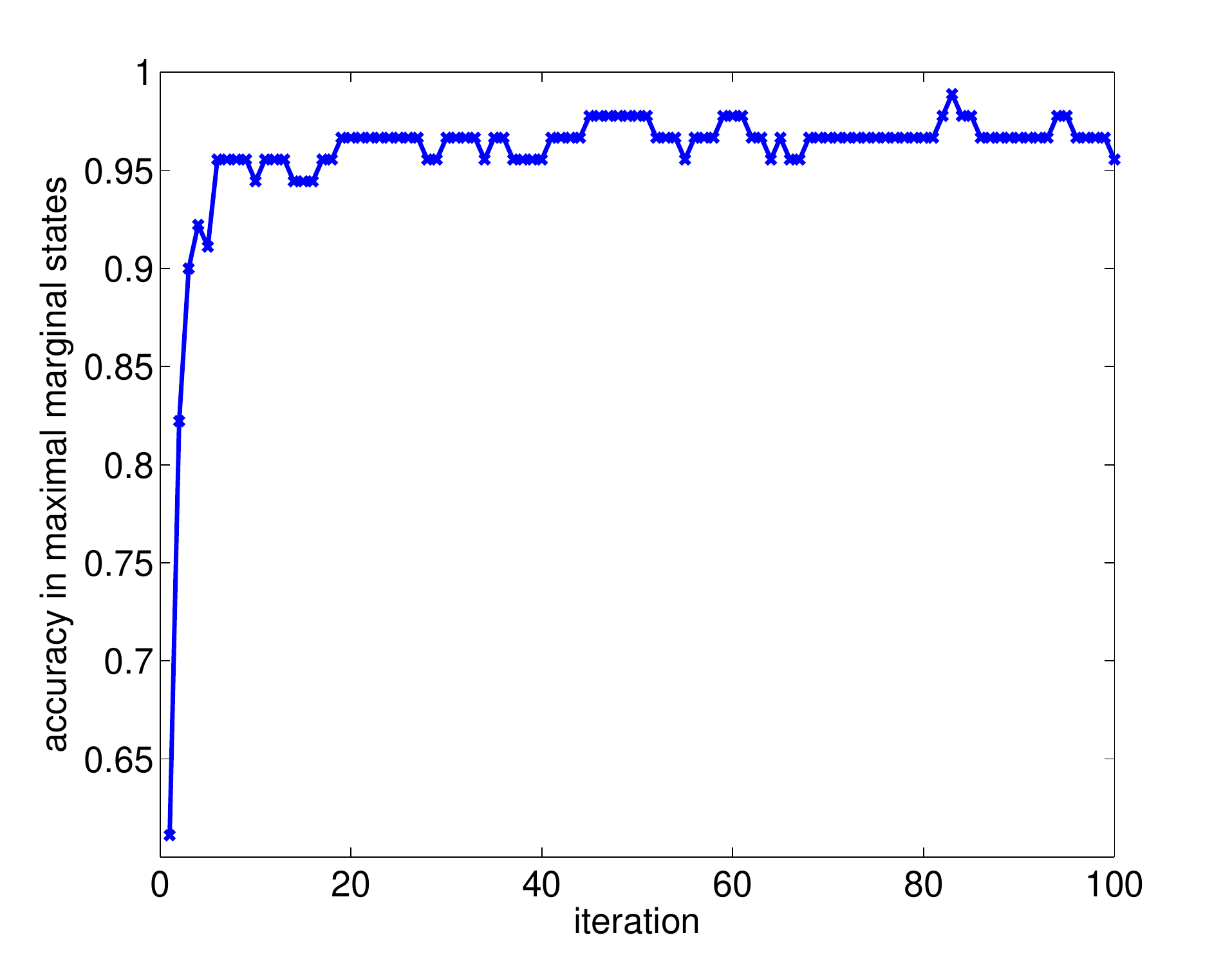} \tabularnewline
\end{tabular}
\par\end{centering}

\caption{Accuracy in term of maximal marginal states.}

\label{fig:state-acc} 
\end{figure}

\textbf{Convergence behaviours for a fixed sequence length:} Figures~\ref{fig:divergences}
and \ref{fig:state-acc} show the three measures using the one-long-run
strategy (up to $5000$ iterations). It appears that both the KL-divergence
and absolute difference continue to decrease, although it occurs at
an extremely slow rate after $500$ iterations. The maximal state
measure is interesting: it quickly reaches the high accuracy after
just $10$ iterations. In other words, the mode of the state marginal
$P(x_{t}^{d}|\o)$ is quickly located although the marginal has yet
to converge.

\begin{figure}[htb]
\begin{centering}
\begin{tabular}{cc}
\includegraphics[width=0.45\linewidth]{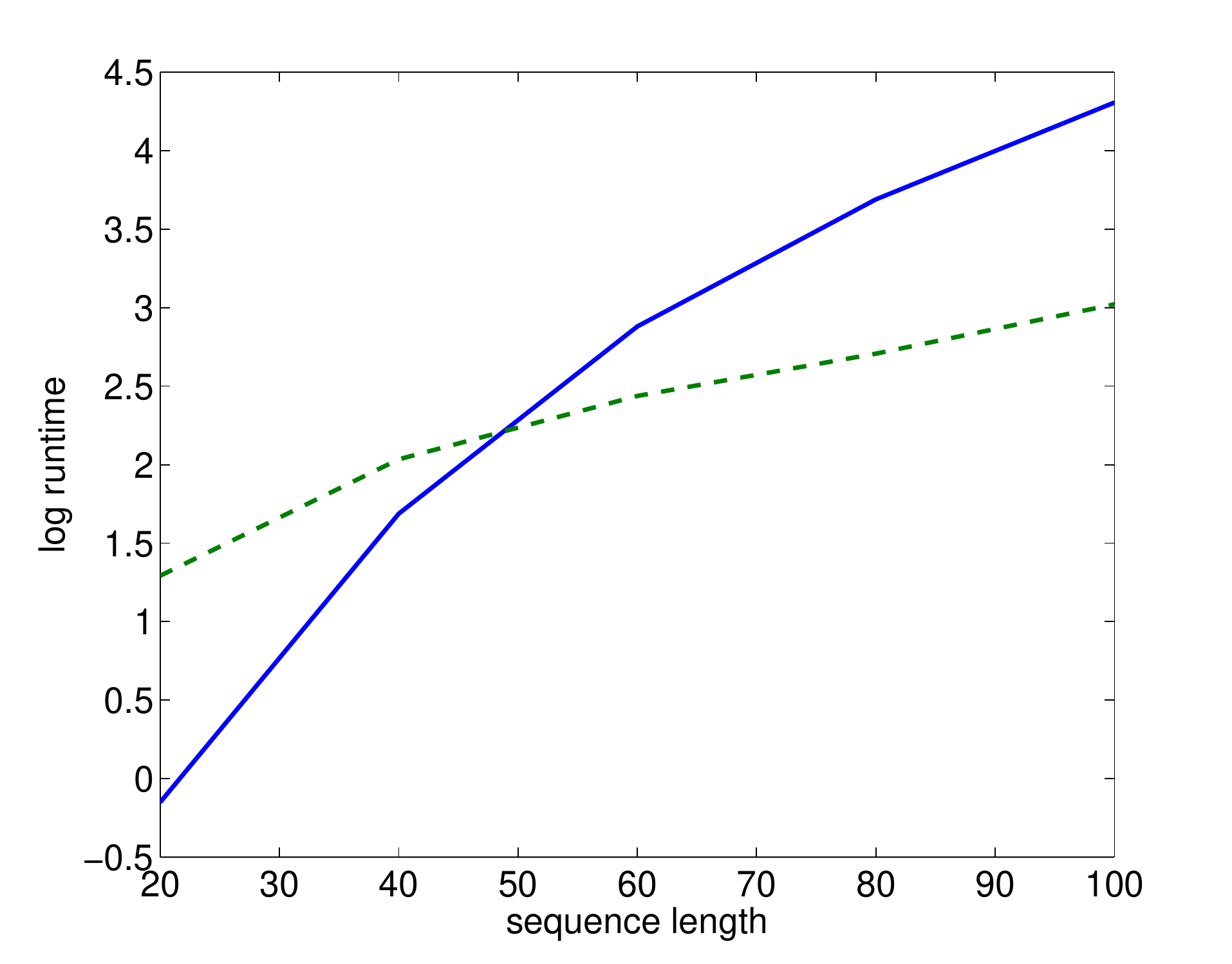}  & \includegraphics[width=0.45\linewidth]{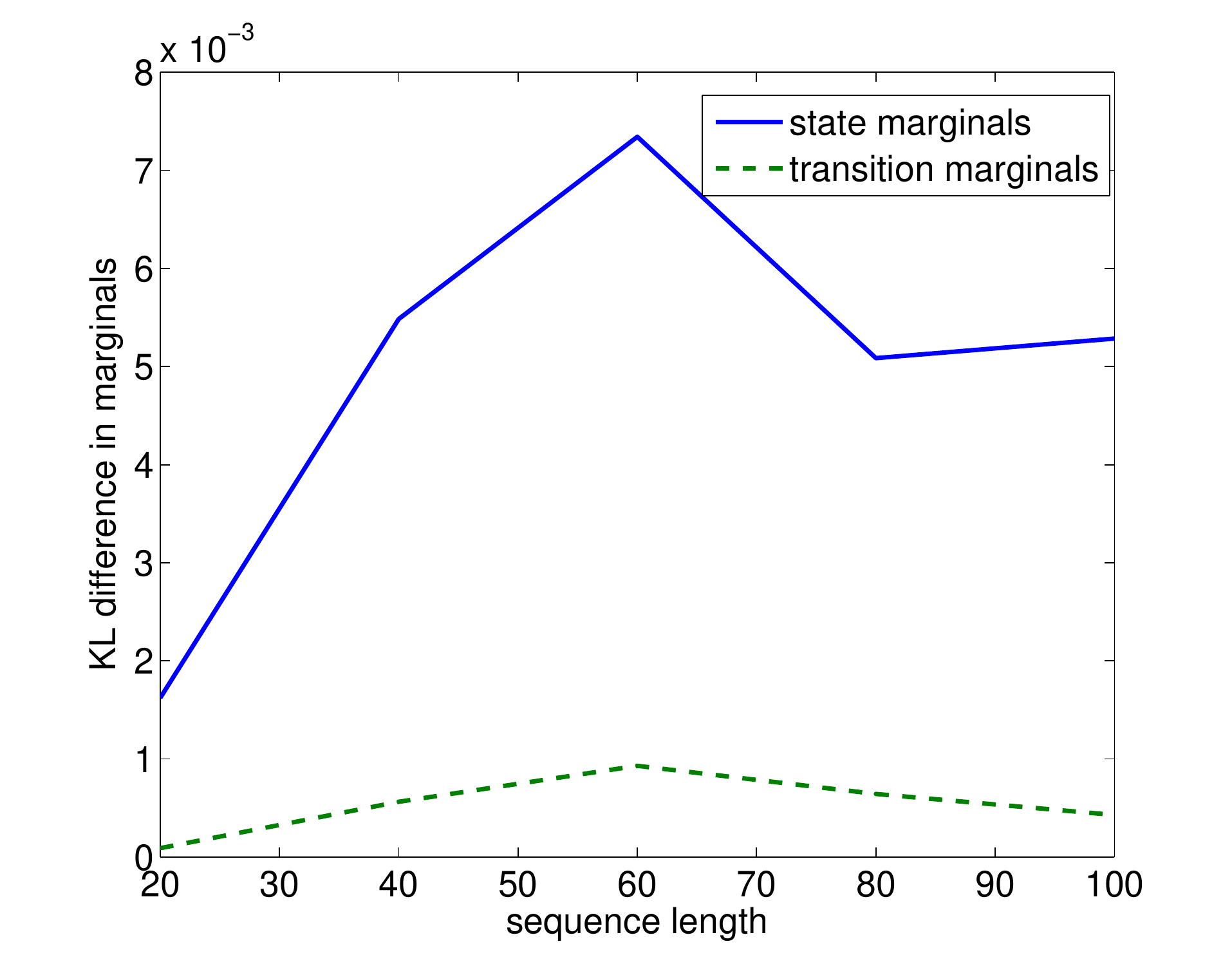} \tabularnewline
(a) log-run time  & (b) KL divergence \tabularnewline
\end{tabular}
\par\end{centering}

\caption{Performance of RBGS in linear run-time w.r.t. sequence length.}

\label{fig:linear} 
\end{figure}

\begin{figure}[htb]
\begin{centering}
\begin{tabular}{cc}
\includegraphics[width=0.45\linewidth]{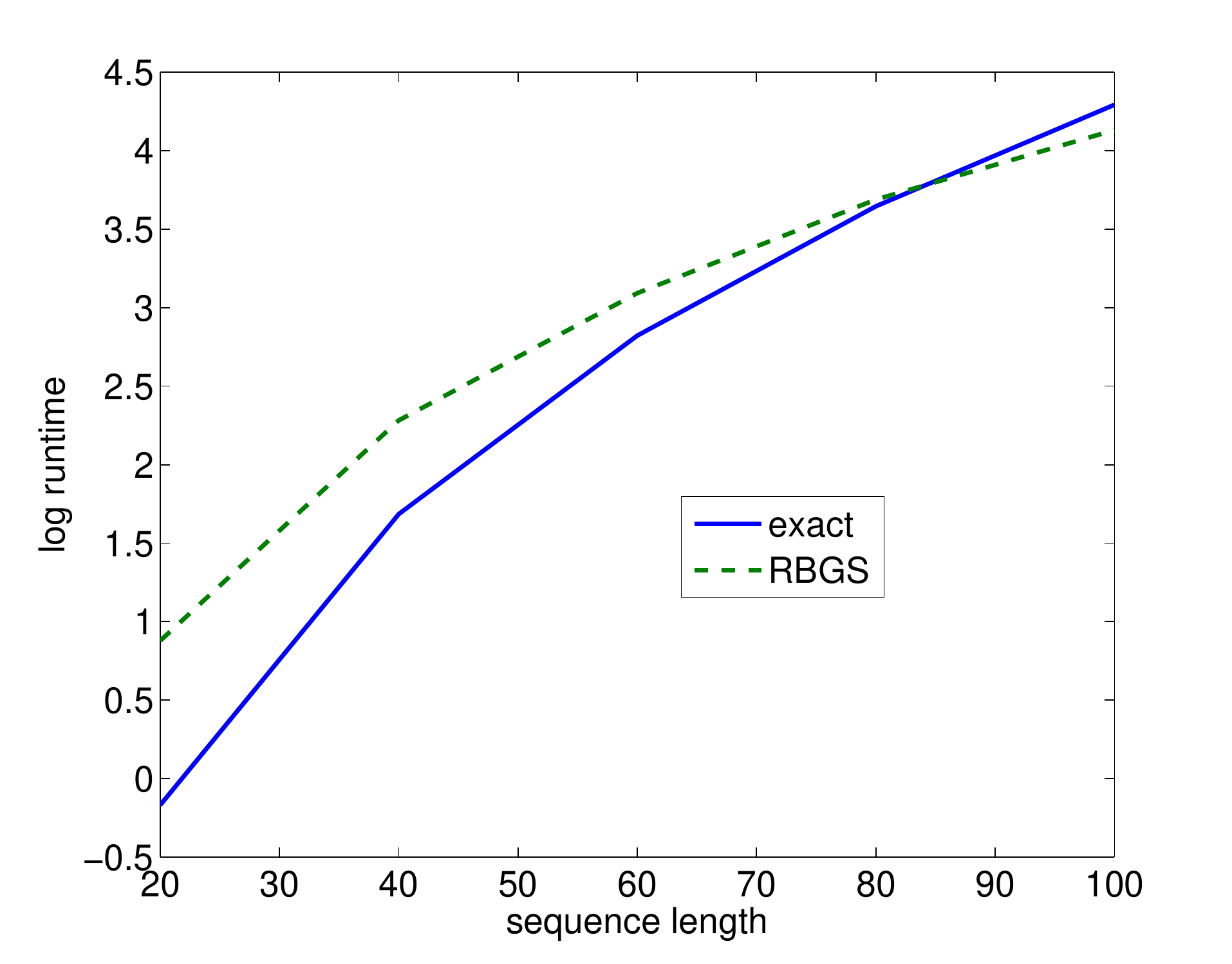}  & \includegraphics[width=0.45\linewidth]{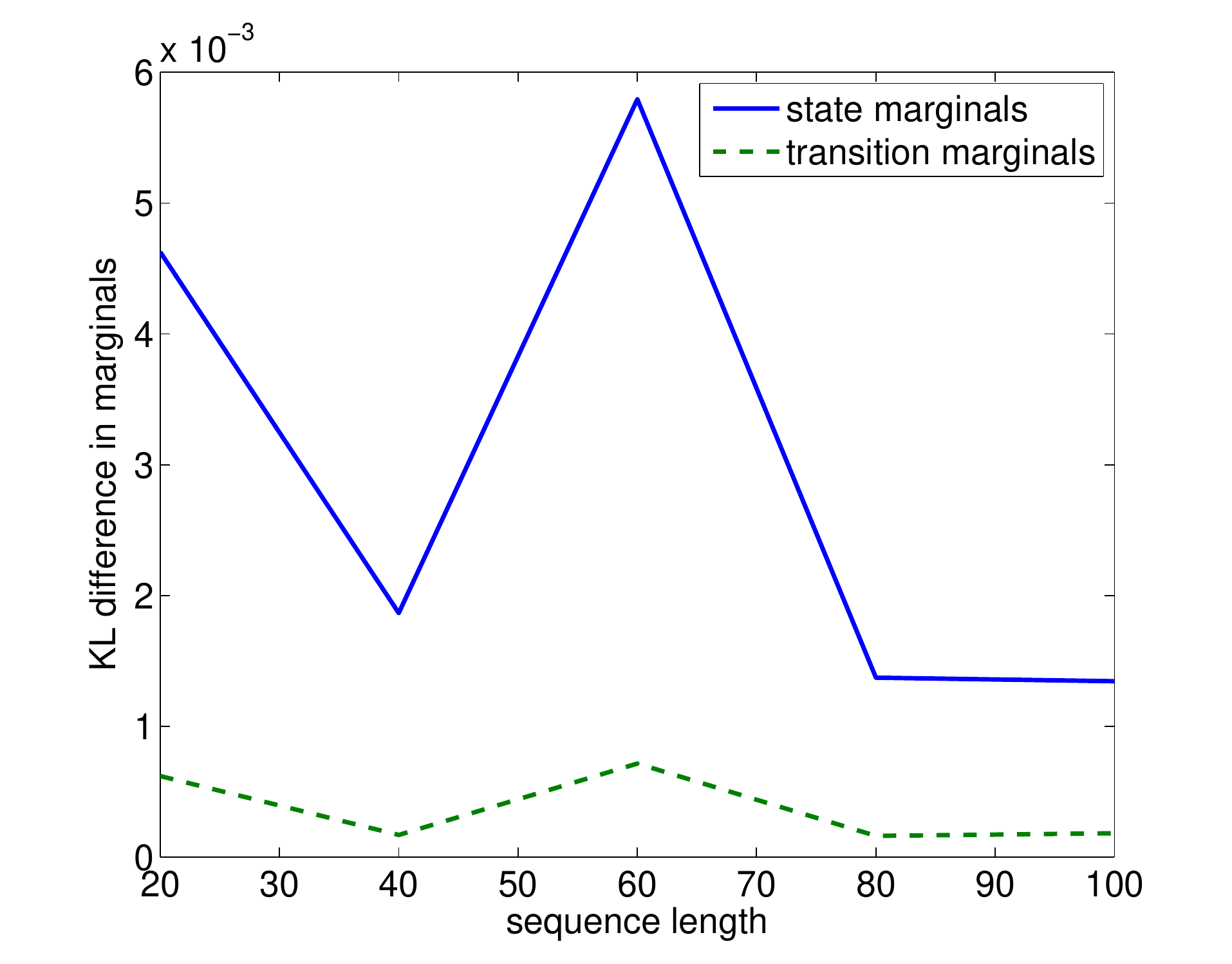} \tabularnewline
(a) log-run time  & (b) KL divergence \tabularnewline
\end{tabular}
\par\end{centering}

\caption{Performance of RBGS in quadratic run-time w.r.t. sequence length.}

\label{fig:quadratic} 
\end{figure}

\begin{figure}[htb]
\begin{centering}
\begin{tabular}{cc}
\includegraphics[width=0.45\linewidth]{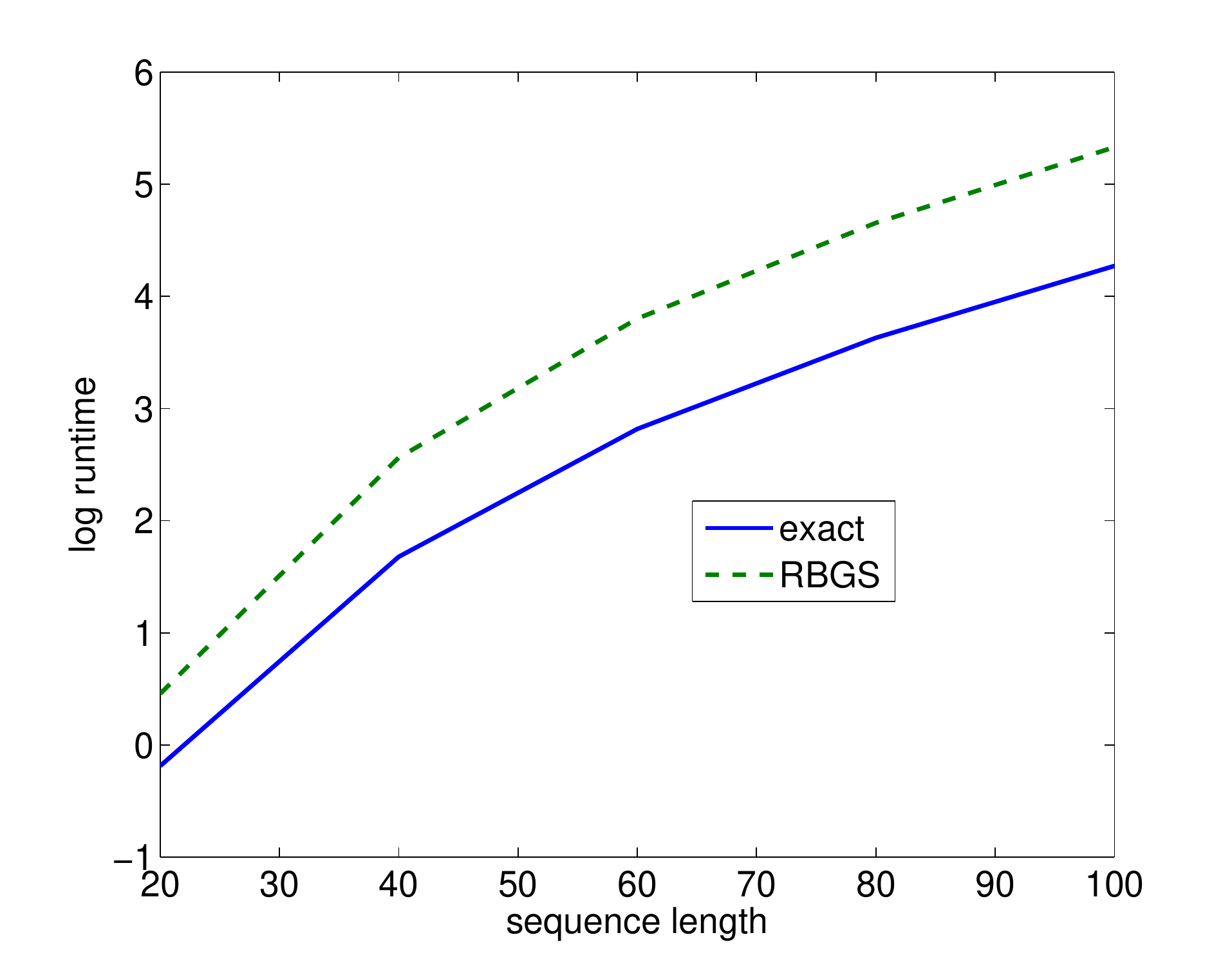}  & \includegraphics[width=0.45\linewidth]{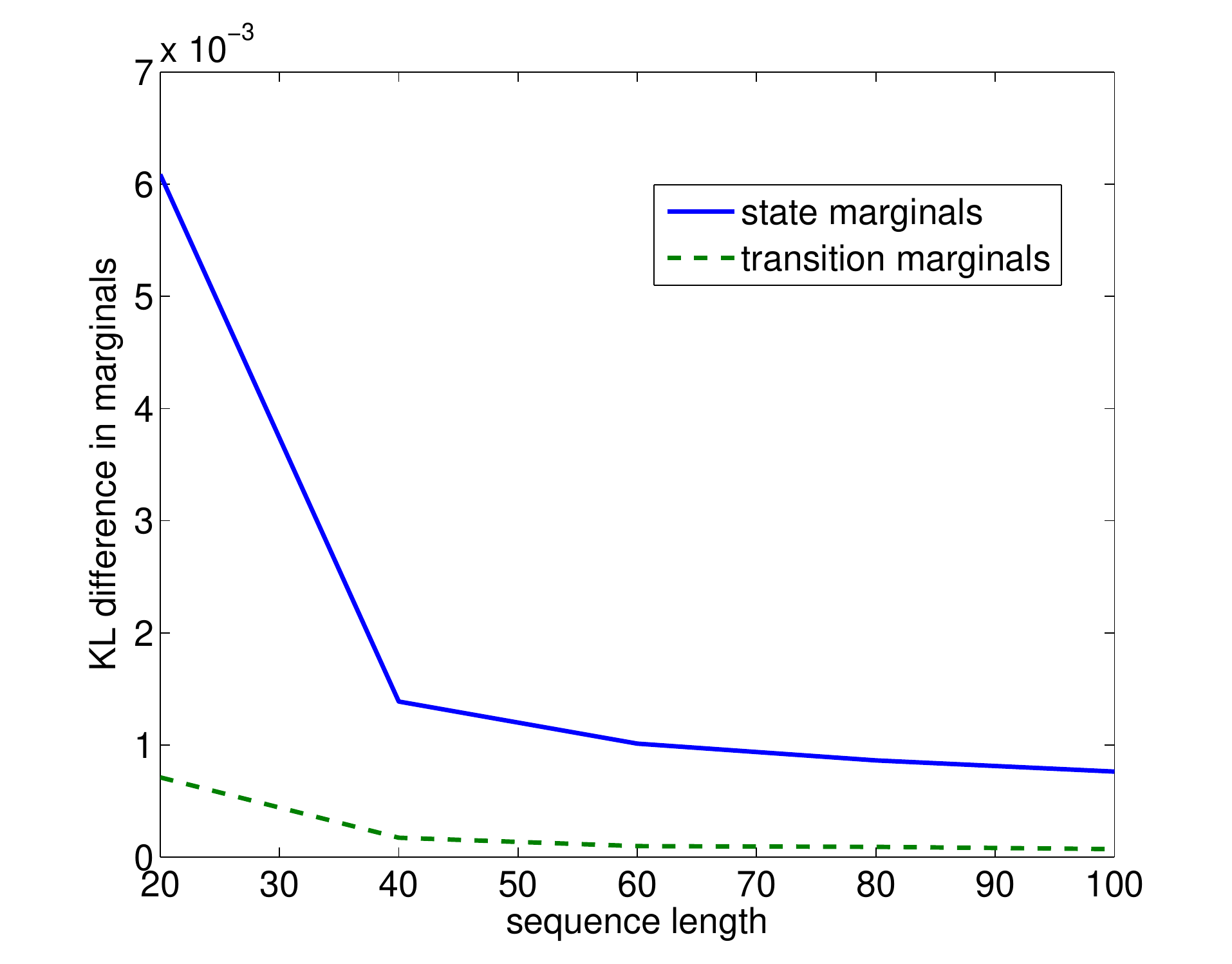} \tabularnewline
(a) log-run time  & (b) KL divergence \tabularnewline
\end{tabular}
\par\end{centering}

\caption{Performance of RBGS in cubic run-time w.r.t. sequence length.}

\label{fig:cubic} 
\end{figure}

\textbf{Convergence as a function of sequence length:} To test how
the RBGS scales, three experiments with different run times are performed: 
\begin{itemize}
\item The number of iterations is fixed at $100$ for all sequence lengths.
This gives us a linear run time in term of $T$. 
\item The number of iterations scales linearly with $T$, and thus the total
cost is $\mathcal{O}(T^{2})$. 
\item The number of iterations scales quadratically with $T$. This costs
$\mathcal{O}(T^{3})$, which means no saving compared to the exact
inference. 
\end{itemize}
The long data is obtained by simply concatenating the original sequences
many times. The sequence lengths are $\{20;40;60;80;100\}$. Figure~\ref{fig:linear}(a,b)
show the performance of the RBGS in linear run-time with respect to
sequence length. Figure~\ref{fig:linear}a depicts the actual run-time
for the exact and RBGS inference. Figure~\ref{fig:linear}b plots
the KL-divergences between marginals obtained by the exact method
and by the RBGS. Performance of the RBGS for quadratic and cubic run-times
is shown in Figure~\ref{fig:quadratic}(a,b) and Figure~\ref{fig:cubic}(a,b),
respectively.

\section{Relations with Other Deep Architectures}

Deep neural architectures (DNA) such as Deep Belief Networks \cite{hinton2006rdd}
and Deep Boltzmann Machines \cite{salakhutdinov2009deep} have recently
re-emerged as a powerful modelling framework which can potentially
discover high-level semantics of the data. The HSCRF shares some similarity
with the DNA in the way that they both use stacking of simpler building
blocks. The purpose is to capture \emph{long-range} dependencies or
\emph{higher-order} correlations which are not directly evident in
the raw data. The building blocks in the HSCRF are the chain-like
Conditional Random Fields, while they are the Restricted Boltzmann
Machines in the DNA. These building blocks are different, and as a
result, the HSCRF is inherently sequential and localised in state
representation, while the DNA was initially defined for non-sequential
data and distributed representation. In general, the distributed representation
is richer as it carries more bits of information given a number of
hidden units. The drawback is that probabilistic inference of RBM
and its stacking is intractable, and thus approximation techniques
such as MCMC and mean-field are often used. Inference in the HSCRF,
on the other hand, is polynomial. For approximation, the MCMC technique
proposed for the HSCRF in this paper exploits the efficiency in localised
state representation so that the Rao-Blackwellisation can be used.

Perhaps the biggest difference between the HSCRF and DNA is the modelling
purpose. More specifically, the HSCRF is mainly designed for discriminative
mapping between the sequential input and the nested states, usually
in a fully supervised fashion. The states often have specific meaning
(e.g. noun-phrase, verb-phrase in sentence modelling). On the other
hand, DNA is for discovering hidden features, whose meanings are often
unknown in advance. Thus, it is generative and unsupervised in nature.
Finally, despite this initial difference, the HSCRF can be readily
modified to become a generative and unsupervised version, such as
the one described in \cite[Ch. 9]{truyen-thesis08}.

Training in HSCRFs can be done simultaneously across all levels, while
for the DNA it is usually carried out in a layer-wise fashion. The
drawback of the layer-wise training is that errors made by the lower
layers often propagate to the higher. Consequently, an extra global
fine tuning step is often employed to correct them.

There have been extensions of the Deep Networks to sequential patterns
such as Temporal Restricted Boltzmann Machines (TRBM) \cite{sutskever2007learning}
and some other variants. The TRBM is built by stacking RBMs both in
depth and time. Another way to build deep sequential model is to feed
the top layer of the Deep Networks into the chain-like CRF, as in
\cite{Do-Artieres-DLSRRA09}.

\section{Conclusion}

We have introduced a novel technique known as Rao-Blackwellised Gibbs
Sampling (RBGS) for approximate inference in the Hierarchical Semi-Markov
Conditional Random Fields. The goal is to avoid both the cubic-time
complexity in the standard Inside-Outside algorithms and the exponential
states in the DBN representation of the HSCRFs. We provide some simulation-based
evaluation of the quality of the RGBS with respect to run time and
sequence length.

This work, however, is still at an early stage and there are promising
directions to follow. First, there are techniques to speed up the
mixing rate of the MCMC sampler. Second, the RBGS can be equipped
with the Contrastive Divergence \cite{Hinton02} for stochastic gradient
learning. And finally, the ideas need to be tested on real, large-scale
applications with arbitrary length and depth.


\begin{thebibliography}{10}

\bibitem{Bui-et-al04}
H.~H. Bui, D.~Q. Phung, and S.~Venkatesh.
\newblock Hierarchical hidden {M}arkov models with general state hierarchy.
\newblock In D.~L. McGuinness and G.~Ferguson, editors, {\em Proceedings of the
  19th National Conference on Artificial Intelligence (AAAI)}, pages 324--329,
  San Jose, CA, Jul 2004.

\bibitem{Bui-et-al02}
H.~H. Bui, S.~Venkatesh, and G.~West.
\newblock Policy recognition in the abstract hidden {M}arkov model.
\newblock {\em Journal of Artificial Intelligence Research}, 17:451--499, 2002.

\bibitem{casella1996rbs}
G.~Casella and C.P. Robert.
\newblock {Rao-Blackwellisation of sampling schemes}.
\newblock {\em Biometrika}, 83(1):81, 1996.

\bibitem{Do-Artieres-DLSRRA09}
Trinh-Minh-Tri Do and Thierry Artieres.
\newblock {Neural conditional random fields}.
\newblock In {\em NIPS Workshop on Deep Learning for Speech Recognition and
  Related Applications}, 2009.

\bibitem{doucet2000rao}
A.~Doucet, N.~de~Freitas, K.~Murphy, and S.~Russell.
\newblock {Rao-Blackwellised particle filtering for dynamic Bayesian networks}.
\newblock In {\em Proceedings of the Sixteenth Conference on Uncertainty in
  Artificial Intelligence}, pages 176--183. Citeseer, 2000.

\bibitem{fine98hierarchical}
Shai Fine, Yoram Singer, and Naftali Tishby.
\newblock The hierarchical hidden {M}arkov model: Analysis and applications.
\newblock {\em Machine Learning}, 32(1):41--62, 1998.

\bibitem{Geman-Geman84}
S.~Geman and D.~Geman.
\newblock Stochastic relaxation, {G}ibbs distributions, and the {B}ayesian
  restoration of images.
\newblock {\em IEEE Transactions on Pattern Analysis and Machine Intelligence
  (PAMI)}, 6(6):721--742, 1984.

\bibitem{Hinton02}
G.E. Hinton.
\newblock Training products of experts by minimizing contrastive divergence.
\newblock {\em Neural Computation}, 14:1771--1800, 2002.

\bibitem{hinton2006rdd}
G.E. Hinton and R.R. Salakhutdinov.
\newblock Reducing the dimensionality of data with neural networks.
\newblock {\em Science}, 313(5786):504--507, 2006.

\bibitem{lafferty01conditional}
J.~Lafferty, A.~McCallum, and F.~Pereira.
\newblock Conditional random fields: Probabilistic models for segmenting and
  labeling sequence data.
\newblock In {\em Proceedings of the International Conference on Machine
  learning (ICML)}, pages 282--289, 2001.

\bibitem{Murphy-Paskin01}
K.~Murphy and M.~Paskin.
\newblock Linear time inference in hierarchical {HMM}s.
\newblock In {\em Advances in Neural Information Processing Systems (NIPS)},
  volume~2, pages 833--840. MIT Press, 2002.

\bibitem{RabinerIEEE-89}
Lawrence~R. Rabiner.
\newblock A tutorial on hidden {M}arkov models and selected applications in
  speech recognition.
\newblock {\em Proceedings of the {IEEE}}, 77(2):257--286, 1989.

\bibitem{salakhutdinov2009deep}
R.~Salakhutdinov and G.~Hinton.
\newblock {Deep Boltzmann Machines}.
\newblock In {\em Proceedings of 20th AISTATS}, volume~5, pages 448--455, 2009.

\bibitem{sarawagi04}
Sunita Sarawagi and William~W. Cohen.
\newblock Semi-{M}arkov conditional random fields for information extraction.
\newblock In Bottou~L Saul~LK, Weiss~Y, editor, {\em Advances in Neural
  Information Processing Systems 17}, pages 1185--1192. MIT Press, Cambridge,
  Massachusetts, 2004.

\bibitem{sutskever2007learning}
I.~Sutskever and G.E. Hinton.
\newblock {Learning multilevel distributed representations for high-dimensional
  sequences}.
\newblock In {\em Proceeding of the Eleventh International Conference on
  Artificial Intelligence and Statistics}, pages 544--551, 2007.

\bibitem{truyen-thesis08}
Tran~The Truyen.
\newblock {\em On Conditional Random Fields: Applications, Feature Selection,
  Parameter Estimation and Hierarchical Modelling}.
\newblock PhD thesis, Curtin University of Technology, 2008.

\bibitem{Truyen:2008a}
T.T. Truyen, D.Q. Phung, H.H. Bui, and S.~Venkatesh.
\newblock {Hierarchical Semi-Markov Conditional Random Fields for recursive
  sequential data}.
\newblock In {\em Twenty-Second Annual Conference on Neural Information
  Processing Systems (NIPS)}, Vancouver, Canada, Dec 2008.

\end{thebibliography}
\end{document}